3D modelling of survey scene from images enhanced with a multi-exposure fusion

Kwok-Leung Chan[1,*], Liping Li[2], Arthur Wing-Tak Leung[3] and Ho-Yin Chan[4]

[1] Department of Electrical Engineering, City University of Hong Kong; itklchan@cityu.edu.hk

[2] Department of Electrical Engineering, City University of Hong Kong; lipingli3-c@my.cityu.edu.hk

[3] Division of Building Science and Technology, City University of Hong Kong; bsawtl@cityu.edu.hk

[4] Department of Electrical Engineering, City University of Hong Kong; hychan656-c@my.cityu.edu.hk

* Correspondence: itklchan@cityu.edu.hk

**Abstract**

In current practice, scene survey is carried out by workers using total stations. The method has high accuracy, but it incurs high costs if continuous monitoring is needed. Techniques based on photogrammetry, with the relatively cheaper digital cameras, have gained wide applications in many fields. Besides point measurement, photogrammetry can also create a three-dimensional (3D) model of the scene. Accurate 3D model reconstruction depends on high quality images. De-graded images will result in large errors in the reconstructed 3D model. In this paper, we pro-pose a method that can be used to improve the visibility of the images, and eventually reduce the errors of the 3D scene model. The idea is inspired by image dehazing. Each original image is first transformed into multiple exposure images by means of gamma-correction operations and adaptive histogram equalization. The transformed images are analyzed by the computation of the local binary patterns. The image is then enhanced, with each pixel generated from the set of transformed image pixels weighted by a function of the local pattern feature and image saturation. Performance evaluation has been performed on benchmark image dehazing datasets. Experimentations have been carried out on outdoor and indoor surveys. Our analysis finds that the method works on different types of degradation that exist in both outdoor and indoor images. When fed into the photogrammetry software, the enhanced images can reconstruct 3D scene models with sub-millimeter mean errors.

## 1. Introduction

An accurate survey of the scene is important for monitoring the ground settlement in construction sites, or assessing the geometric accuracy and structural health of a building. For instance, tunneling work may result in ground surface displacement that could potentially affect adjacent properties. Spaceborne synthetic aperture radar interferometry (In-SAR) can be used for ground settlement monitoring over a wide spatial area. Liu et al. [1] employed the time series InSAR technique to investigate the temporal and spatial deformation of land reclamation. For close-range ground settlement monitoring, workers use a total station to measure heights at some points of the survey scene. The method has high accuracy but it incurs high costs due to continuous monitoring. The structure of buildings and infrastructures could be deformed due to external factors such as winds, or earthquakes. Batur et al. [2] performed a structural health assessment of historical structures using terrestrial laser scanning (TLS) technology. Ge et al. [3] employed TLS to detect changes to ground surfaces and buildings. The monitoring results are comparably with global positioning system (GPS) measurements. Mikrut [4] employed laser scanner and also photogrammetry model obtained from digital images in the reconstruction of historical buildings.

Structure from motion (SfM) photogrammetry reconstructs 3D model using overlap-ping images captured from different viewpoints. It offers a low-cost solution with the use of consumer grade digital cameras. SfM can achieve camera calibration without the need for control points. It is widely employed for non-contact remote monitoring. Moses et al. [5] presented a survey on methods used for measuring the changes of rock surface such as TLS and SfM. Cullen et al. [6] compared SfM and a contact method using erosion meter. Erosion meter can produce very accurate measure, but the number of measurement points is limited. On contrary, SfM can produce dense 3D measure, and sub-millimeter accuracy is possible with simple planar surfaces.

Techniques based on photogrammetry, with the relatively cheaper digital cameras, have gained wide applications in many fields. Besides point measurement, photogrammetry can also create high quality three-dimensional (3D) models of the scene. Moreover, it considerably reduces the amount of labor required in the field for similar work. Chian and Yang [7] employed photogrammetry for ground settlement monitoring. With multiple images and the photogrammetry software, 3D coordinates of survey points are estimated with accuracy comparable with measurements from the total station up to millimeters. Baiocchi et al. [8] also employed photogrammetry for monitoring landfill settlement. Images are captured by a digital camera mounted on the unmanned aerial vehicle (UAV). The accuracy of the 3D model is in centimeter, which is close to the laser scanning method. Łabędź et al. [9] compared four image histogram adjustment methods for 3D model reconstruction by photogrammetry.

Photogrammetry estimates the 3D coordinates of points on an object using two or more images taken from different positions. The technique is called triangulation. On each image, common points are located. For each of these points, a ray (line of sight) is formed between the point and the camera position. If the camera location and aiming direction are known, the 3D (XYZ) coordinates of the point can be estimated by the intersection of these rays. The accuracy would be enhanced if more images showing the target point is available (i.e. higher percentage of overlaps between images). If there are control points present in the image, the image coordinates can be converted to the real coordinates (mm) – geo-referencing.

The performance of 3D model reconstruction can be affected by the quality of the images. For instance, if images are of low contrast, the number of matched points between neighboring images will be reduced. Figure 1 shows one original survey image, the image enhanced by our proposed method, and the corresponding histograms. The enhanced image contains more pixels in both the low and high intensities. Figure 2(a) shows the matching of two original images captured in adjacent positions. The images are matched using image feature SURF. The numbers of detected points in the left image and the right image are 5,523 and 6,374 respectively. Figure 2(b) shows the matching of the two corresponding enhanced images. The numbers of detected points in the left image and the right image are 6,745 and 7,919 respectively. The enhanced images, with better visibility, have more points detected. It can be seen that the enhanced images can have more matched points (see the top left corner and the right side of the picture).

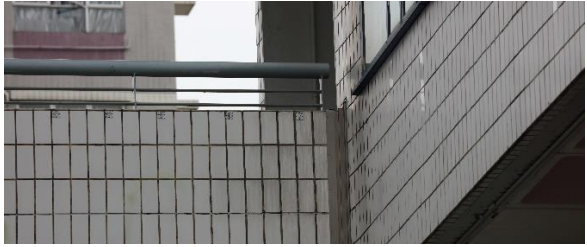 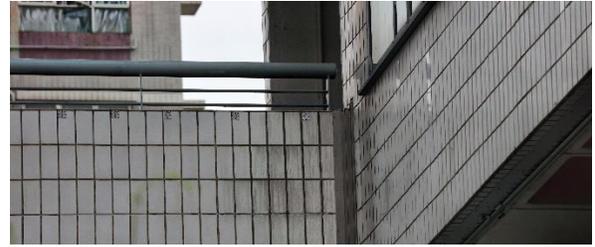

(a)                                        (b)

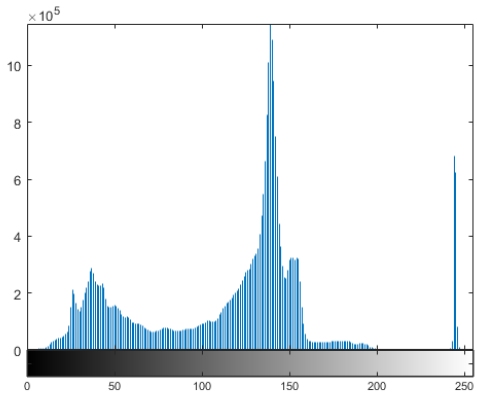 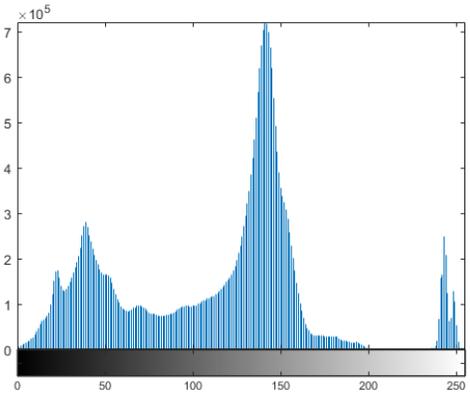

(c)                                        (d)

Figure 1. (a) Original survey image, (b) enhanced image, (c) histogram of the original image, (d) histogram of the enhanced image.

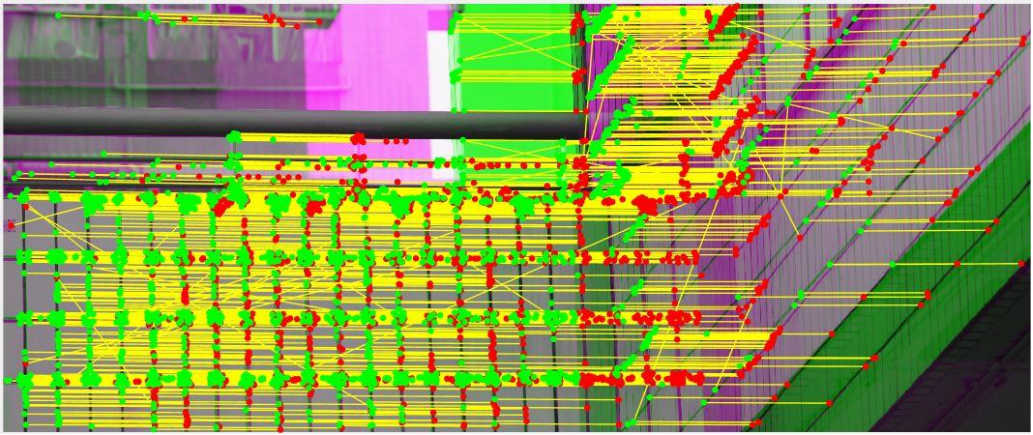

(a)

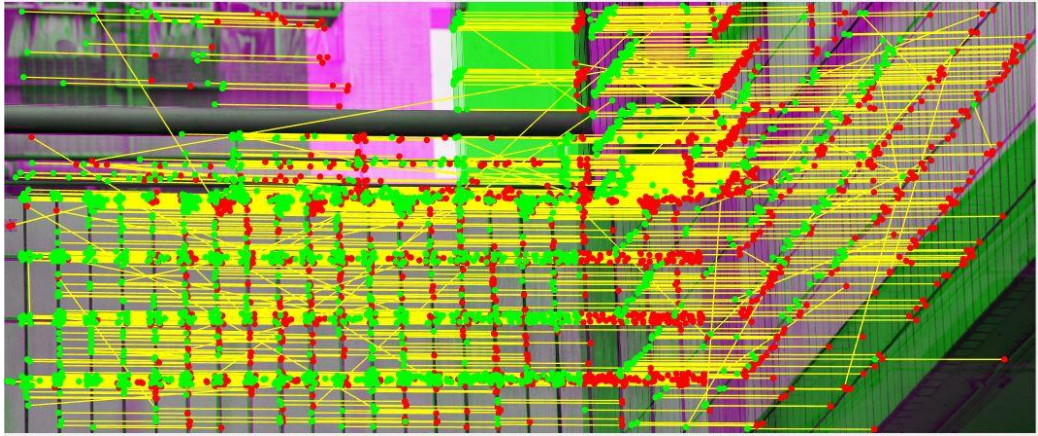

(b)

Figure 2. (a) Matching of two original images, (b) matching of two enhanced images.

Image acquisition systems may produce degraded images. For instances, bad weather (haze, fog) can cause images captured outdoor with low visibility, indoor images can be degraded by the existence of smoke. Eliminating haze and fog in images can significantly improve the visibility of the outdoor scene. When the image colors are distorted by airlight, image dehazing can help compensate for the color shift. The dehazed image is more visually pleasing. Many computer vision applications assume that the input image is the scene radiance. Haze removal, with the recovery of image radiance, ensures reliable features extracted from images. Image dehazing has become popular research in computer vision. As presented in a recent survey [10], the number of publications on this topic has increased every year.

In this paper, we propose an image enhancement method. Survey images are pre-processed in order to eliminate the degradation. The enhanced images, when fed into the photogrammetry software, can reconstruct 3D scene models with sub-millimeter accuracy. Our contributions are as follows:

- We develop a low cost, high accuracy 3D scene model reconstruction system based on photogrammetry. With the enhancement of the survey images, the accuracy of the measurement points is in sub-millimeters, which is comparable to the total station.

- We propose a novel method for enhancing the survey images. Inspired by the idea of image dehazing, each image is transformed into multiple exposure images. A fusion strategy is used to generate the enhanced image. Performance evaluation and comparative analysis have been carried out on publicly available image dehazing datasets.

- To the best of our knowledge, our proposed method is the first to utilize the local image pattern to characterize the degree of haziness. The multiple exposure images, based on the pattern features, are integrated to generate the dehazed image.

- To investigate the usefulness of the image enhancement method, experimentations have been performed on outdoor and indoor surveys. The enhanced images, when fed into the photogrammetry software, can reconstruct 3D scene models with sub-millimeter mean errors. The quantitative and visual results show that our image enhancement method significantly improves the reconstructed 3D model over the original images, and also outperforms other state-of-the-art image dehazing methods.

The organization of this paper is as follows. The related researches are reviewed in Section 2. We focus on various algorithms proposed for enhancing the visibility of the image. Section 3 describes the equipment being used and the complete process for reconstructing the 3D model of the survey scene. Section 4 explains in detail our proposed image enhancement method. We evaluate our proposed method and compare its performance with various well-known and state-of-the-art algorithms. Section 5 presents the experimental results and comparative analysis. Finally, we conclude our work and suggest some future work in Section 6.

## 2. Related work

We group the related work on image dehazing into three parts: enhancement of survey image mainly with image processing techniques, image dehazing methods based on prior information or physical model, and benchmarks for image dehazing research.

### 2.1. Image enhancement for survey image

Images captured in the outdoor survey may suffer from bad weather conditions such as haze and fog. Images captured in an indoor environment may suffer from poor visibility or dark colors due to illumination variations. Many computer vision applications, e.g. remote sensing and surveillance, demand high quality images. Image enhancement techniques have been developed aiming to restore the visibility of degraded images. Chaudhry et al. [11] proposed a framework to remove haze in outdoor images based on median filtering and Laplacian filtering. Guo et al. [12] observed the correlation between adjacent bands in multi-spectral remote sensing images. They proposed a haze removal method via estimation of haze thickness, atmospheric light, and transmission value. Li et al. [13] also analyzed the image haziness within a local patch. They proposed the sphere model improved dark channel prior (DCP) for transmission estimation which is suitable for thin haze, as well as uneven and thick haze images. With such a model, they developed a haze removal method based on homomorphic filtering.

Salazar-Colores et al. [14] combined DCP with morphological and Gaussian filters in the image dehazing framework. Zhang et al. [15] used the saliency detection method to locate bright regions. Those regions are excluded from the estimation of transmission and airlight. Dharejo et al. [16] proposed a simple color correction algorithm to improve the color and contrast of hazy images. First, the color shift is rectified by the piecewise linear transformation. Second, the sharpness of the image is improved by the optical contrast method. Liu et al. [17] classified image dehazing methods into three categories: image enhancement based on image processing algorithms with handcrafted features, image dehazing algorithms based on prior information or physical model, and deep learning image dehazing networks.

### 2.2. Physical model-based image dehazing

Wang et al. [18] grouped image dehazing methods into three categories according to the processing techniques: enhancement based, fusion based and restoration based. Enhancement based methods utilize image processing algorithms to improve the visibility of the image. Fusion based methods generate the dehazed image from multiple input images. Restoration based methods adopt a degradation model and restore a high-quality image by reversing the degradation processes. He et al. [19] proposed the DCP to remove haze in a single image. From observation, they found that haze-free outdoor images contain low intensity in local patches. Transmission and airlight are estimated. The scene radiance is computed based on the haze imaging model. Since then, many

DCP based methods have been proposed. Xiao et al. [20] performed sky region segmentation. The estimated airlight is more accurate than DCP.

Li et al. [21] grouped image dehazing methods into four categories according to the inputs: multiple-image based, polarizing filter based, known depth based, and single-image based. Galdran [22] developed an image dehazing technique based on multiple exposure images. With the prior knowledge of hazy images and the physical model, the pixelwise hazy-free color is estimated by a multi-scale Laplacian blending scheme.

Recently, computer vision has advanced rapidly through the use of deep learning. In contrast to algorithms based on handcrafted features, deep learning is a machine learning based on learning data representations. Babu and Venkatram [10] presented a comprehensive review on state-of-the-art haze removal techniques. Recently, there are more image dehazing techniques based on machine learning and deep learning approaches. Li et al. [23] proposed the All-in-One Dehazing Network (AOD-Net). A light-weight CNN learns the transmission map, which is then fed into the atmospheric scattering model for haze-free image generation. Jiao et al. [24] proposed an end-to-end learnable dehazing network to jointly estimate and refine the transmission map. The dehazed image is then generated by the physical model. Although deep learning models can be trained to produce very good results with benchmark datasets, their performance can deteriorate significantly on unseen images.

### 2.3. Image dehazing datasets

Image dehazing has become an important computer vision topic in recent years. Publicly available datasets are created to facilitate image dehazing research. One difficulty is that the acquisition of ground truth images, e.g. haze-free images in the outdoor environment, is a tedious task. One solution is to generate real haze in a controlled indoor environment with the use of a professional haze generating machine. Ancuti et al. [25] adopted this method and captured clear and hazy images in the same scene at the same conditions. Another method is to simulate the hazy conditions. Tarel et al. simulated fog and created two synthetic datasets, Foggy Road Image DAtabase (FRIDA) [26] and FRI-DA2 [27]. FRIDA contains 90 images of synthetic road scenes. The larger FRIDA2 contains 330 synthetic images exhibiting homogeneous and heterogeneous fog. Zhao et al. [28] created the benchmark dataset, BeDDE, which is probably the first dataset containing both real outdoor clear and hazy images. Since the hazy and the corresponding clear images were often captured in slightly different positions, the authors also manually defined the region of interest (ROI) for each image pair that can be used for a relevant quantitative evaluation.

Comparison of dehazing methods can be performed by subjective and objective evaluations. There are three classes of image dehazing evaluation schemes. First, helpers are recruited to judge the quality of dehazed images. It may be difficult to demand the helpers to evaluate a large number of images. Also, the judgment is subjective and there may be contradictions among the helpers. Second, dehazed images are evaluated quantitatively with the no-reference image quality assessment (NR-IQA) metrics [29], [30]. The advantage of this scheme is that there is no need for the corresponding clear image in the computation of the numerical measures. However, the NR-IQA metrics may be less reliable than other measures that are computed with reference to the corresponding haze-free image. Third, dehazed images are evaluated quantitatively with the full reference image quality assessment (FR-IQA) metrics. This scheme is probably the most prevalent one. By comparing the dehazed image and the clear image, numerical measures such as the peak-signal-to-noise ratio (PSNR) and structural similarity index measure (SSIM) [31] can be computed.

## 3. 3D model reconstruction

During the survey experiment, a number of markers are placed in the scene. 3D coordinates of the survey markers, also called control points, are measured by the total station. At the same time, images of the survey scene are captured by two digital cameras. With the control points and the images input to the photogrammetry software, the 3D model of the survey scene is reconstructed and coordinates of the model are converted to the real coordinates in millimeter via the process of geo-referencing. In this section, we first illustrate the process of measuring the survey markers. Then, we describe the image acquisition and 3D model reconstruction.

### 3.1. Distance measurement for survey markers

The surveying experiments aim to establish the coordinates of a set of markers for subsequent photogrammetry processing. As shown in Figure 3, reflective sheets are used as survey markers that stick on a solid background or are fixed on supporting frames. A total station, an electronic distance measurement (EDM) equipment, is used to measure the distance and coordinates of the markers from the observation point (OP). It has a sighting accuracy of 2 mm at a distance of 100 m when using a reflective sheet. The markers are positioned to reflect the light beam to the total station at an angle closer to the right angle to obtain accurate measurements. The total station is assumed to be set on a "real" traverse station as the OP. The coordinates of the OP are based on a Traverse Station near City University. The coordinates of the markers are measured and recorded in sequence. In each survey, 4 to 6 sets of data were measured and recorded to assure the accuracy of the measurements. The 360-prism will also be used as a target to verify the coordinates when necessary. The average coordinates of the markers are converted into HK1980 Grid Coordinates with reference to the OP coordinates.

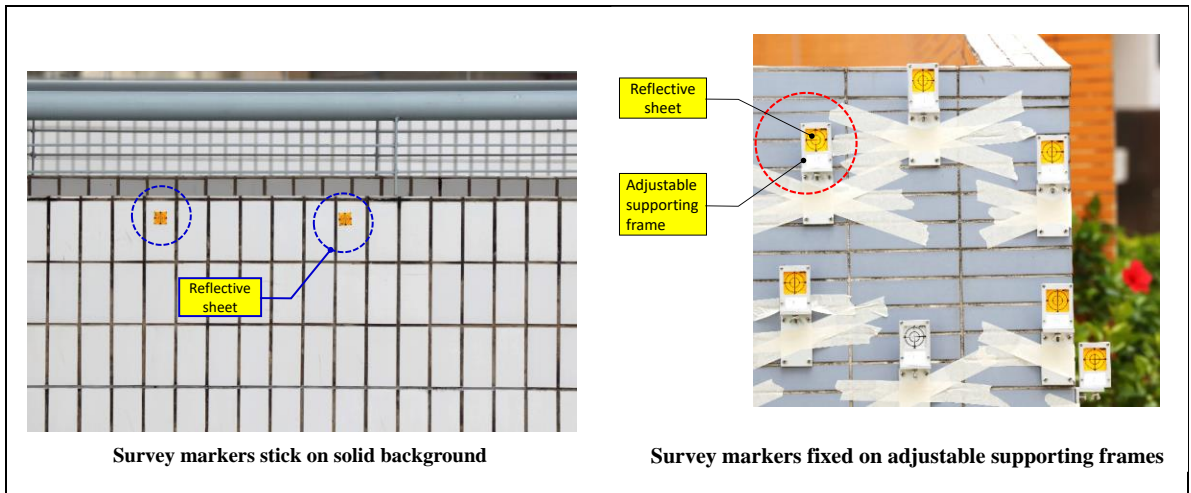

**Survey markers stick on solid background**   **Survey markers fixed on adjustable supporting frames**

Figure 3. Survey markers.

### 3.2. Image acquisition and 3D model reconstruction

We use two digital cameras (resolution 8,688 x 5,792 pixels, 70-300 mm lens) for image acquisition. Each camera is set at a certain distance from the survey scene. The two cameras are separated at a baseline distance from each other. Both cameras are set at approximately the same height. During image acquisition, we focus on the survey markers. Precautions are taken to avoid image blur caused by camera vibrations. Neighboring images should have a 70% overlap or more. In each image, the control points are marked and labelled manually. Images and the control points are input to the photogrammetry software Pix4D [32]. Pix4D first performs camera calibration, followed by the generation of point cloud and 3D surface model. It also produces a quality report, which shows the errors of the control points and other information.

## 4. Image enhancement

We observe that the accuracy of the reconstructed 3D scene model can be lowered due to image degradation. Therefore, we propose an enhancement method to pre-process the images. Inspired by the idea of image dehazing, our method can improve the visibility and contrast of the images, and eventually reduce the errors of the 3D model. We explain our image dehazing algorithm in section 4.1. The key characteristic of our method is to estimate the haziness of the image based on local pattern analysis. The local texture pattern is described in section 4.2.

### 4.1. Image dehazing

Figure 4 shows the image enhancement framework. The original image is first transformed into multiple exposure images. The transformed images are analyzed by the computation of local texture pattern and local saturation. Extraction of local textural features will be described in detail in the next sub-section. We propose to characterize the degree of haziness by a weighted function of the local texture pattern feature and local image saturation. The dehazed image is then generated based on the weighted function and the set of transformed image pixels at the corresponding position.

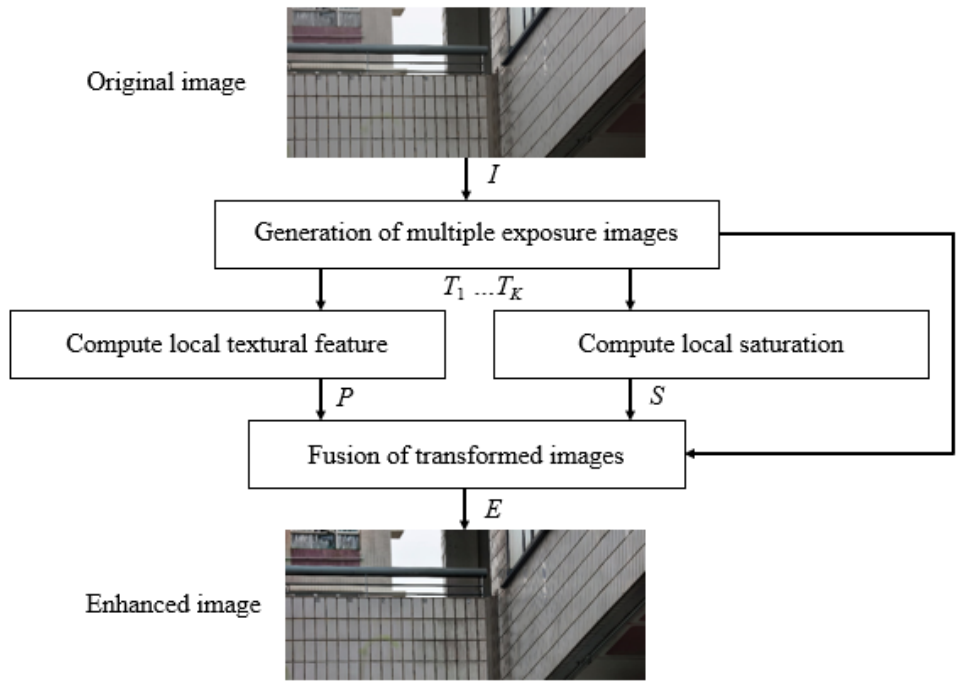

Figure 4. Image enhancement framework.

We adopt two methods to transform the hazy image to 5 images with different exposures. The first method to acquire 4 images with different exposures is to transform the original image by gamma correction. The second method is to synthesize the fifth contrast-enhanced image by Contrast-Limited Adaptive Histogram Equalization (CLAHE). Gamma correction modifies the image intensities $I$ with the power transform function

$$T = I^\gamma \tag{1}$$

where $\gamma$ is the gamma factor. With $\gamma > 1$, brighter image intensities are mapped to a wider range, while darker image intensities are mapped to a narrower range. That means the variations in the bright image region are less noticeable, while the details in the dark region are more perceptible. We propose an adaptive scheme that can generate a suitable set of gamma correction images. If the average intensity of the image is less than a threshold value, the gamma factors are set as (1.2, 1.4, 1.6 and 1.8). Otherwise, the gamma factors are set as (2, 3, 4 and 5). In that sense, a darker image will be transformed with a moderate intensity range compression. On the contrary, a brighter image, transformed with a larger gamma factor, will allocate lesser intensity values to bright pixels (i.e., wider range). In such case, more intensity values can be allocated to dark pixels.

Another method to improve the contrast of the image is histogram equalization. Instead of transforming the image intensities globally, CLAHE partitions the image into a number of regions and equalizes the histogram of the regions one by one. The boundary between neighboring regions is eliminated with bilinear interpolation. The amount of contrast enhancement is governed by a single parameter called clip limit. A higher value of clip limit will result in more contrast in the image region.

Pixelwise image saturation $S(x)$ is computed by the standard deviation across all color channels

$$S(x) = \sqrt{\sum_{c \in \{R,G,B\}} \left( T^c(x) - \frac{T^R(x) + T^G(x) + T^B(x)}{3} \right)^2} \tag{2}$$

where $T(x)$ is the transformed image intensity at location $x$. A larger standard deviation will weigh more in the generation of dehazed image pixels. Therefore, $S(x)$ will tend to utilize rich colors from the set of multiple exposure images in synthesizing the enhanced image.

The enhanced image $E(x)$ is generated by a fusion of the set of transformed images $T(x)$. To avoid the artifacts produced by the integration process, each transformed image is represented by a multi-scale structure via the construction of a Laplacian pyramid. Each level of the pyramid $L$ is computed as the difference between the transformed image at that scale and the upsampled version of the transformed image at the next lower resolution scale

$$L^i = T^i - \uparrow (T^{i+1}) \tag{3}$$

where $i$ is the scale index, and $\uparrow$ is the upsampling operation. The weight is represented by a multi-scale structure via the construction of Gaussian pyramid $H$. Each weight map is computed as the product of local textural feature $P$ (which will be explained in the following sub-section), and $S$.

$$H^i = P^i \cdot S^i \tag{4}$$

Finally, the enhanced image is generated by the fusion

$$E(x) = \sum_{i=1}^{N} \sum_{k=1}^{K} H_k^i(x) L_k^i(x) \tag{5}$$

where $N$ is the number of levels in the multi-scale structure, and $K$ is the number of multiple exposure images.

*4.2. Estimation of haziness by local texture*

Another factor that can characterize the degree of image haziness is the local texture pattern feature. Local image patterns and the features estimated from the spatial domain are effective image texture representations. Image pattern, with multiple pixels, is more informative than a single pixel. Local image pattern has been adopted for other applications. For instance, Liao et al. [33] proposed the scale invariant local ternary pattern for moving objects detection in video. St-Charles et al. [34] used local binary similarity pattern for background modeling.

In this paper, we propose the textural pattern for characterizing the local contrast of the image. It has two advantages. First, the coded pattern is perceptually correlated with the local contrast. Second, the textural features, which are computed from all color channels, are better than other grey-scale textural features.

Assume the size of the local pattern is 3 x 3 pixels. Each pixel of the pattern (except the center pixel) is coded with three labels *l* by the following equation

$$l_n = \begin{cases} 0, & CI_l \leq c_n \leq CI_u \\ +1, & c_n > CI_u \\ -1, & c_n < CI_l \end{cases}, 1 \leq n \leq 8 \qquad (6)$$

where $n$ is the position index, $c \in \{R, G, B\}$ is the color component value. The label is determined by comparing the color component value with a range called confidence interval defined by the lower bound $CI_l$ and upper bound $CI_u$. The confidence interval is specified by the following equations

$$CI_l = c_0 - \alpha c_0 \qquad (7)$$

$$CI_u = c_0 + \beta c_0 \qquad (8)$$

where $c_0$ is the color component value of the center pixel. $\alpha$ and $\beta$ are defined by

$$\alpha = \frac{10^{\frac{D_{min}}{20}} - 1}{10^{\frac{D_{min}}{20}}} \qquad (9)$$

$$\beta = 10^{\frac{D_{min}}{20}} - 1 \qquad (10)$$

where $D_{min}$ is the minimal distinguishable difference, which is set equal to 0.5 dB. Figure 5 shows two 3 x 3 pixels image patterns (the numbers are the color component values) and the corresponding coded patterns. The code labels characterize the visually perceptible difference between neighboring pixels and the center pixel.

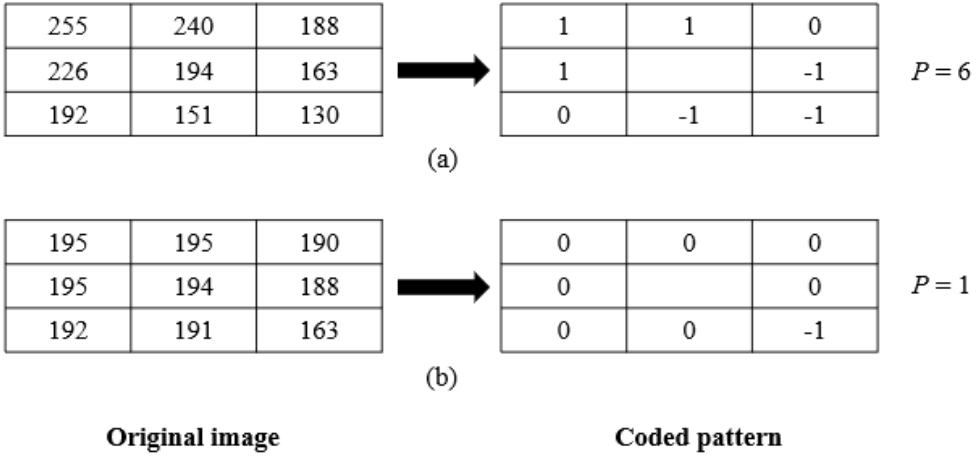

|        |        | (a)    |
| ------ | ------ | ------ |

Original image                          Coded pattern

Figure 5. Examples of coding of the local image patterns with corresponding feature values.

The coded pattern is then concisely represented as a textural feature of the center pixel $P$, which is one parameter for computation of the weight map (see equation 4). We devise a simple and effective conversion method with the following equation

$$P = \sum_{c \in \{R,G,B\}} |l_-^c| + \sum_{c \in \{R,G,B\}} l_+^c \tag{11}$$

where $l_-$ and $l_+$ correspond to the negative label value and positive label value respectively. Finally, $P$ is normalized by the size of the local pattern. The local pattern with higher contrast (i.e., large differences between neighboring pixels and center pixel), e.g. Figure 5 (a), will have a higher $P$. As a result, the center pixel can contribute more to the generation of the enhanced image. On the contrary, a pattern with more zero labels, e.g. Figure 5 (b), will have a lower $P$. That center pixel should contribute less in the weight map.

## 5. Experimental results and discussion

We first evaluate our image enhancement method on synthetic and real hazy image datasets. FRIDA2 [27] contains synthetic images exhibiting homogeneous and heterogeneous fog. BeDDE [28] contains real outdoor hazy images. Comparative analysis is per-formed with various image dehazing algorithms. These reference methods are deterministic (DCP [18], AMEF [20]), and deep learning-based (AOD-Net [23]). Three Full-Reference (FR) Image Quality Assessment (IQA) metrics are adopted. Mean-Squared Error (MSE) and Peak Signal-to-Noise Ratio (PSNR) are commonly used fidelity measures. They are expressed as

$$MSE(R,E) = \frac{1}{MN} \sum_{i=0}^{M-1} \sum_{j=0}^{N-1} (R(i,j) - E(i,j))^2 \tag{12}$$

$$PSNR(R,E) = 10 log_{10} \left( \frac{(L-1)^2}{MSE(R,E)} \right) \tag{13}$$

where $R$ and $E$ denote the reference and enhanced images respectively, $M$ and $N$ denote the height and width of the image respectively, and $L$ is the maximum range of the image intensity values. Structural Similarity (SSIM), measuring the similarity between the reference image and enhanced image, is expressed as

$$SSIM(R,E) = \frac{(2\mu_R\mu_E + C_1)(2\sigma_{RE} + C_2)}{(\mu_R^2 + \mu_E^2 + C_1)(\sigma_R^2 + \sigma_E^2 + C_2)} \tag{13}$$

where $\mu$ and $\sigma^2$ are the mean and variance of the image respectively, $\sigma_{RE}$ is the covariance of reference image and enhanced image, $C_1 = (k_1(L-1))^2$, $C_2 = (k_2(L-1))^2$, $k_1 = 0.01$, and $k_2 = 0.03$.

For the survey experiments, we compare all methods on the performance of 3D model reconstruction. Mean errors of 3D coordinates of survey markers (with reference to the 3D coordinates measured by the total station) are adopted as quantitative measures.

### 5.1. Performance evaluation on image dehazing datasets

Tables 1 and 2 show the quantitative results of the methods on FRIDA2 and BeDDE respectively. The best result is highlighted in red, while the second-best result is highlighted in blue. Figure 6 shows some visual results on FRIDA2. The first row shows the original image degraded by cloudy homogeneous fog, the clear reference image, and the enhancement results. Similarly, the second row shows the cloudy heterogeneous fog degradation. Part of each image (see the red box) is enlarged. All the enlarged views of the road surface are shown in Figure 7. Figure 8 shows some visual results on BeDDE. Closed-ups of a region are shown in Figure 9.

Table 1. Quantitative results on FRIDA2 dataset.

|  | MSE | PSNR | SSIM |
|---|---|---|---|
| DCP | 0.0898 | 11.0320 | 0.5552 |
| AMEF | **<span style="color:red">0.0627</span>** | **<span style="color:blue">12.3641</span>** | 0.7017 |
| AOD-Net | 0.0680 | 12.0928 | **<span style="color:blue">0.7043</span>** |
| Proposed new method | **<span style="color:blue">0.0631</span>** | **<span style="color:red">12.3817</span>** | **<span style="color:red">0.7273</span>** |

Table 2. Quantitative results on BeDDE dataset.

|  | MSE | PSNR | SSIM |
|---|---|---|---|
| DCP | 0.0395 | 15.1639 | 0.6085 |
| AMEF | **<span style="color:blue">0.0116</span>** | **<span style="color:blue">20.0023</span>** | **<span style="color:blue">0.7325</span>** |
| AOD-Net | 0.0368 | 15.4217 | 0.6027 |
| Proposed new method | **<span style="color:red">0.0075</span>** | **<span style="color:red">22.5224</span>** | **<span style="color:red">0.8252</span>** |

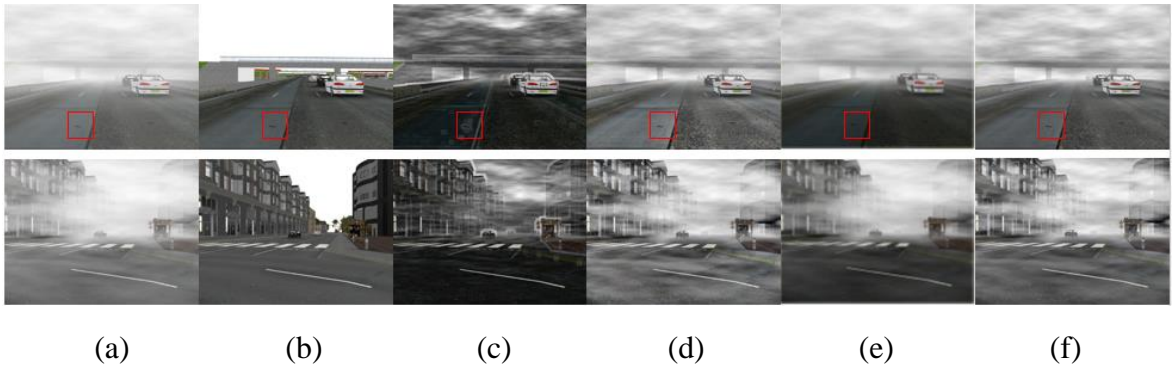

|   (a)   |   (b)   |   (c)   |   (d)   |   (e)   |   (f)   |

Figure 6. Visual results on FRIDA2 dataset: (a) original image, (b) ground truth, (c) DCP, (d) AMEF, (e) AOD-Net, (f) proposed new method.

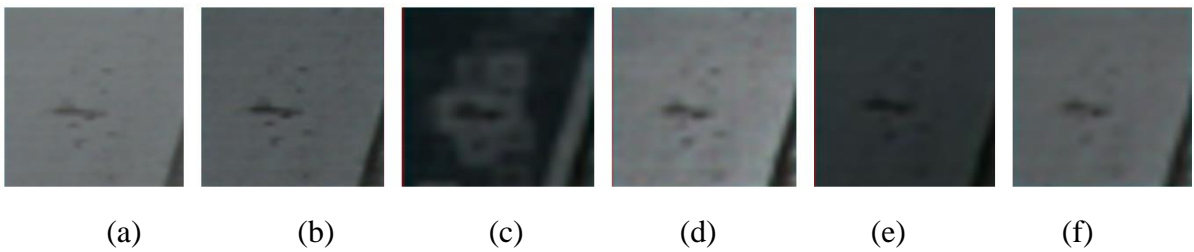

|   (a)   |   (b)   |   (c)   |   (d)   |   (e)   |   (f)   |

Figure 7. Enlarged views of results on FRIDA2: (a) original image, (b) ground truth, (c) DCP, (d) AMEF, (e) AOD-Net, (f) proposed new method.

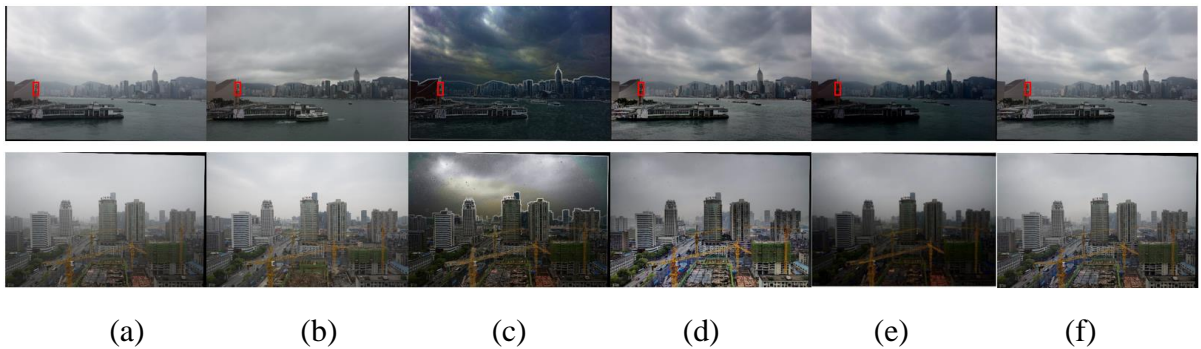

|   (a)   |   (b)   |   (c)   |   (d)   |   (e)   |   (f)   |

Figure 8. Visual results on BeDDE dataset: (a) original image, (b) ground truth, (c) DCP, (d) AMEF, (e) AOD-Net, (f) proposed new method.

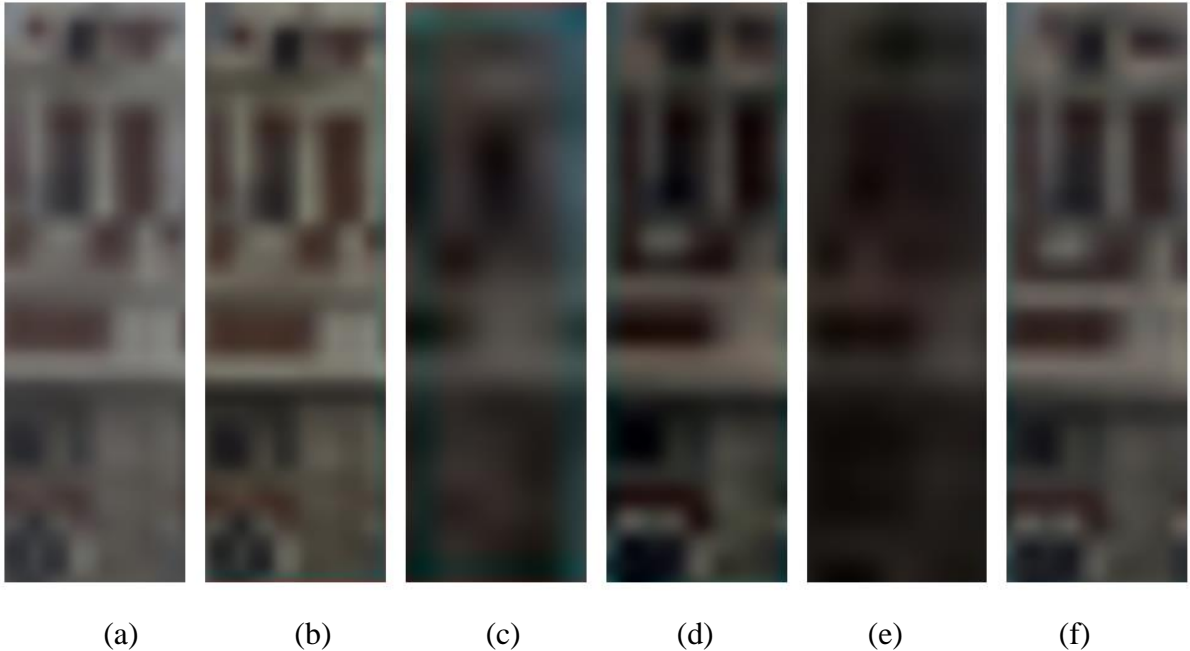

|   (a)   |   (b)   |   (c)   |   (d)   |   (e)   |   (f)   |

Figure 9. Enlarged views of results on BeDDE: (a) original image, (b) ground truth, (c) DCP, (d) AMEF, (e) AOD-Net, (f) proposed new method.

AMEF and proposed new method have better performance than DCP and AOD-Net in terms of all quantitative measures. As shown in the visual results, DCP generates a very dark image with distorted colors. AOD-Net also generates darker image and the texture is blur. AMEF generates accurate texture but the image is brighter than the reference (see Figure 7), or the texture is not as clear as our method (see Figure 9). Our method achieves a good balance in terms of the overall intensity and the detailed texture in the synthesized image.

### 5.2. Outdoor and indoor surveys

Figure 10 shows images of three surveys with the survey markers numbered. Scene 1 is the link bridge between two buildings. The distance between the cameras and the bridge is about 32 m. 30 images were captured. Scene 2 is the link bridge (at a lower level than scene 1). 28 images were captured. Scene 3 is the concourse inside a building. The distance between the cameras and the desk is about 29 m. 25 images were captured. The main challenge is the co-existence of the dark desk and the shiny escalator behind. Tables 3 and 4 show the mean errors and RMS errors respectively of the 3D coordinates of reconstructed survey markers in the first outdoor survey experiment. Tables 5 and 6 show the mean errors and RMS errors respectively of the 3D coordinates of reconstructed survey markers in the second outdoor survey experiment. Tables 7 and 8 show the mean errors and RMS errors respectively of the 3D coordinates of reconstructed survey markers in the indoor survey experiment. The original images generate 3D models with large mean errors, especially for the indoor survey. The 3D models can be improved with images enhanced by DCP and AMEF. However, as will be shown in the following, DCP generates very dark texture map. Both AOD-Net and proposed new method can generate 3D models with sub-millimeter accuracy. In the three surveys, our method can achieve either the best or the second-best accuracy.

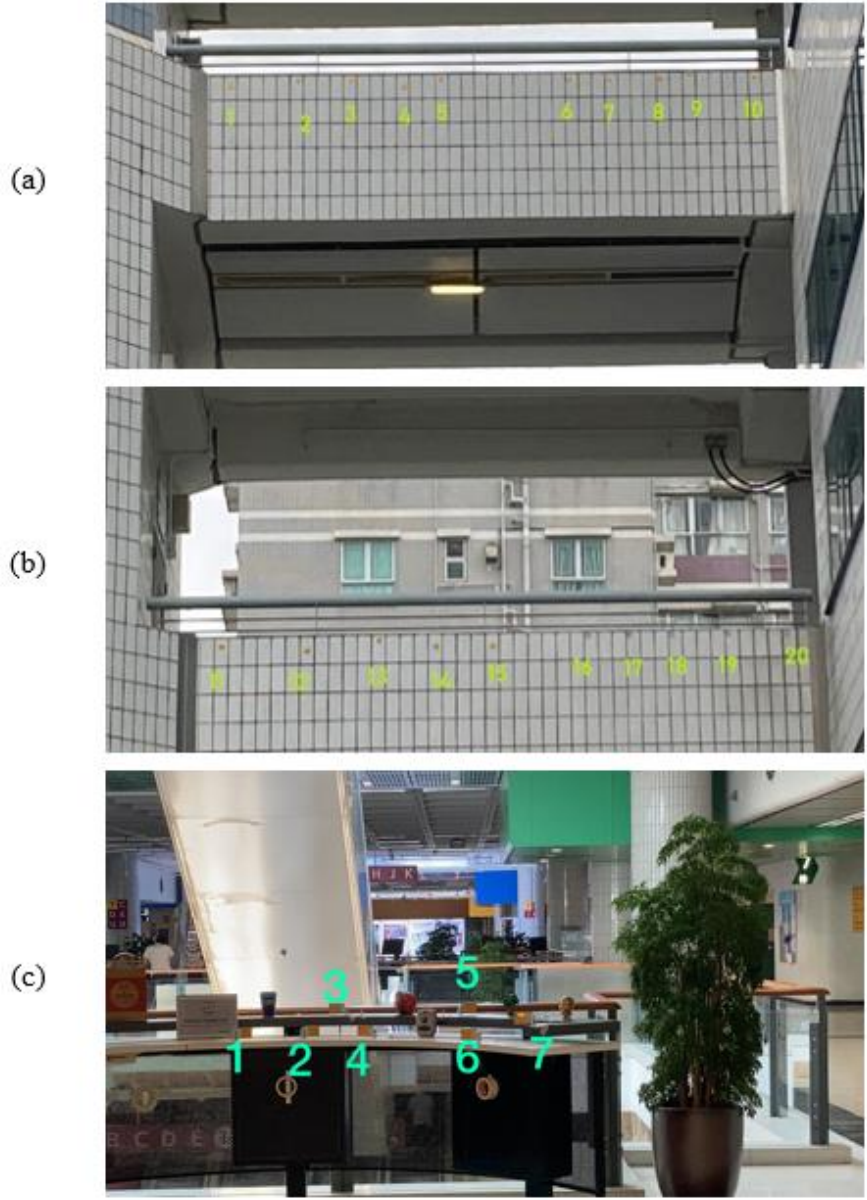

Figure 10. Images of survey experiments: (a) scene 1, (b) scene 2, (c) scene 3.

Table 3. Mean errors (in mm) of outdoor survey 1.

|  | **X** | **Y** | **Z** |
|---|---|---|---|
| Original image | 0.029 | 0.097 | 0.063 |
| DCP | -0.017 | -0.036 | -0.015 |
| AMEF | 0.007 | 0.061 | **0.002** |
| AOD-Net | **0.005** | **0.008** | **-0.000** |
| Proposed new method | **0.002** | **0.017** | **-0.000** |

Table 4. RMS errors (in mm) of outdoor survey 1.

|  | **X** | **Y** | **Z** |
|---|---|---|---|
| Original image | **1.494** | 17.588 | 10.403 |
| DCP | 5.875 | **12.446** | 4.615 |
| AMEF | **1.557** | 19.600 | 2.858 |
| AOD-Net | 2.626 | **12.128** | **2.743** |
| Proposed new method | **1.557** | 14.548 | **2.831** |

Table 5. Mean errors (in mm) of outdoor survey 2.

|  | **X** | **Y** | **Z** |
|---|---|---|---|
| Original image | -0.771 | 0.551 | -0.090 |
| DCP | 1.737 | -1.761 | -1.451 |
| AMEF | **0.007** | 0.155 | **0.005** |
| AOD-Net | **-0.007** | **-0.073** | 0.031 |
| Proposed new method | **0.004** | **0.024** | **0.000** |

Table 6. RMS errors (in mm) of outdoor survey 2.

|  | **X** | **Y** | **Z** |
|---|---|---|---|
| Original image | 22.411 | 31.852 | 43.476 |
| DCP | 6.937 | **5.756** | **5.086** |
| AMEF | 6.606 | **7.196** | 9.184 |
| AOD-Net | **5.780** | 12.803 | **2.603** |
| Proposed new method | **5.611** | 9.346 | 9.064 |

Table 7. Mean errors (in mm) of indoor survey.

|  | **X** | **Y** | **Z** |
|---|---|---|---|
| Original image | -17.219 | -128.054 | 19.119 |
| DCP | 0.558 | -2.348 | -2.896 |
| AMEF | -0.301 | 0.392 | 0.132 |
| AOD-Net | **0.109** | **0.014** | **-0.019** |
| Proposed new method | **-0.070** | **-0.037** | **0.084** |

Table 8. RMS errors (in mm) of indoor survey.

|  | **X** | **Y** | **Z** |
|---|---|---|---|
| Original image | 44.588 | 392.522 | 77.046 |
| DCP | **36.278** | 110.716 | **62.018** |
| AMEF | **28.306** | **63.993** | 70.496 |
| AOD-Net | 60.364 | **39.880** | 67.957 |
| Proposed new method | 52.036 | 222.195 | **16.146** |

Figure 11 shows the point clouds of the first outdoor survey. Figures 12 and 13 show the front views and top views of the 3D mesh models respectively. The original images have fewer matched points than enhanced images. Therefore, the accuracy of the 3D model is lower than those reconstructed by enhanced images (see the distorted straight lines in Figure 12 (a)). The 3D models generated by DCP and AOD-Net have very dark texture maps which are not photorealistic. The reconstructed surface of the wall is also rough. The model generated by AMEF has some defects (see the right side of the wall in Figure 12 (c)). The proposed new method generates the flat surface of the wall (Figure 13 (e)) with a visually pleasing texture map (Figure 12 (e)).

Figure 14 shows the point clouds of the second outdoor survey. Figures 15 and 16 show the front views and top views of the 3D mesh models respectively. The original images cannot match well and very small parts of the surface can be reconstructed. DCP and AOD-Net generate very dark texture maps. The wall reconstructed by AOD-Net is rough (Figure 16 (d)). AMEF and proposed new method can generate accurate 3D model. Comparing Figure 16 (c) and (e), our method can reconstruct a fairly flat surface of the wall than AMEF.

Figure 17 shows the point clouds of the indoor survey. Figures 18 and 19 show the front views and top views of the 3D mesh models respectively. The scene is complicated as both very dark (desk) and very bright (escalator) regions exist in each image. The original images, DCP, and AOD-Net cannot generate good 3D models. AMEF (Figure 19 (c)) and proposed new method (Figure 19 (e)) can reconstruct very accurate curved surface of the desk.

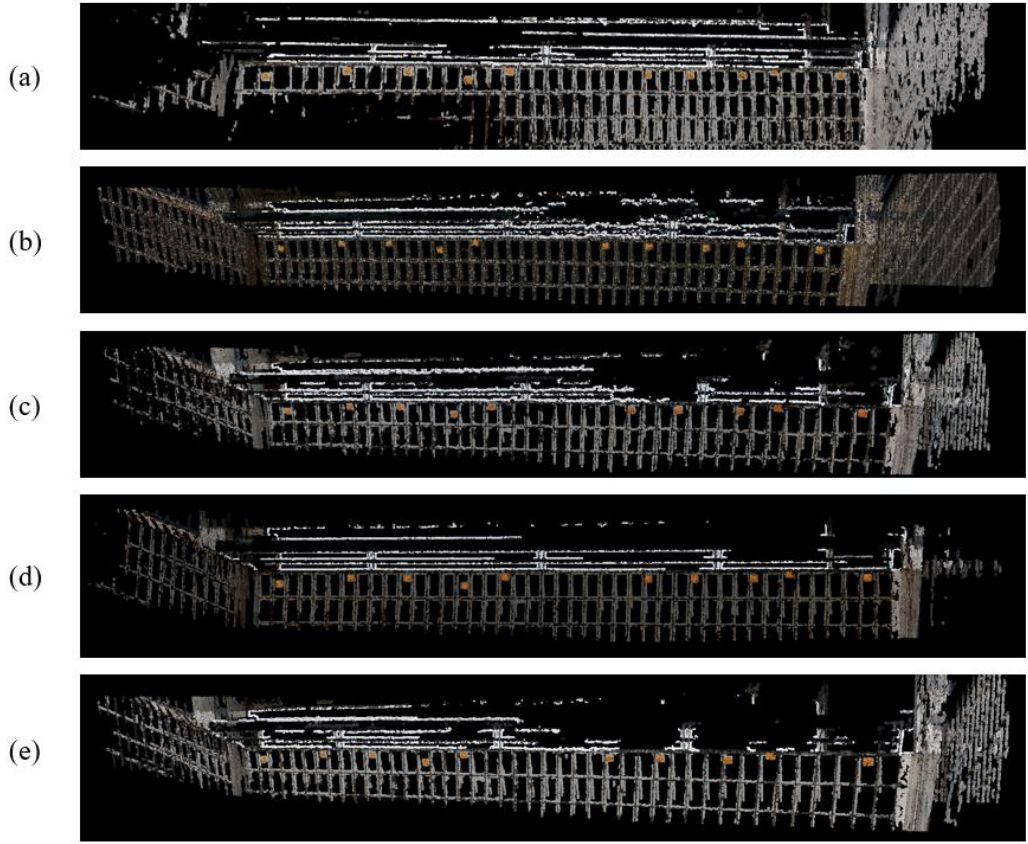

Figure 11. Point clouds of survey experiment 1 reconstructed from: (a) original images, (b) images enhanced by DCP, (c) images enhanced by AMEF, (d) images enhanced by AOD-Net, (e) images enhanced by proposed new method.

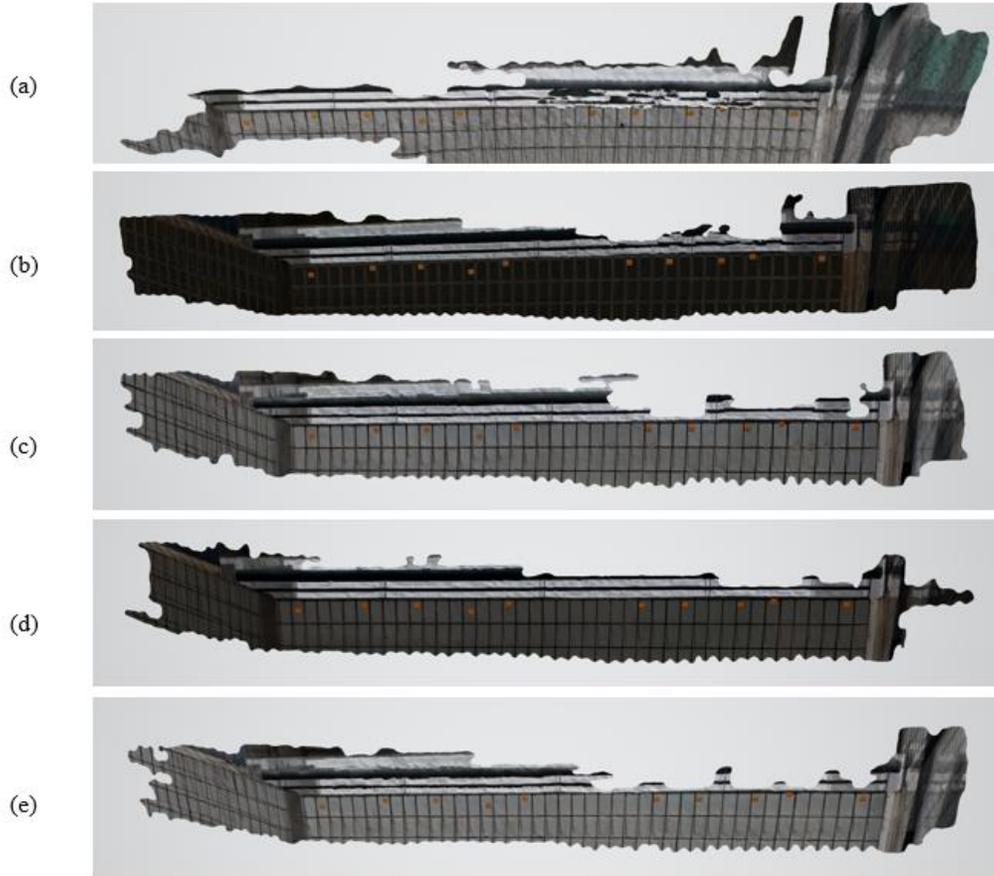

Figure 12. 3D mesh models (front views) of survey experiment 1 reconstructed from: (a) original images, (b) images enhanced by DCP, (c) images enhanced by AMEF, (d) images enhanced by AOD-Net, (e) images enhanced by proposed new method.

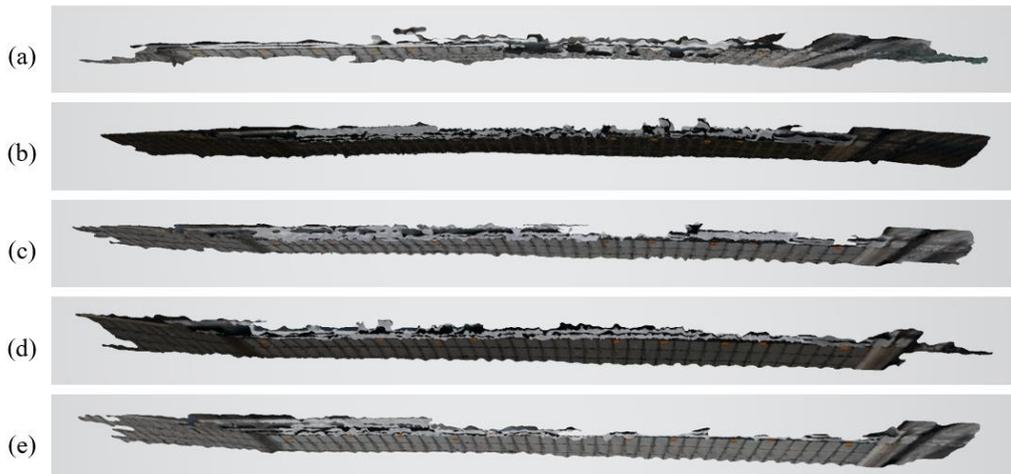

Figure 13. 3D mesh models (top views) of survey experiment 1 reconstructed from: (a) original images, (b) images enhanced by DCP, (c) images enhanced by AMEF, (d) images enhanced by AOD-Net, (e) images enhanced by proposed new method.

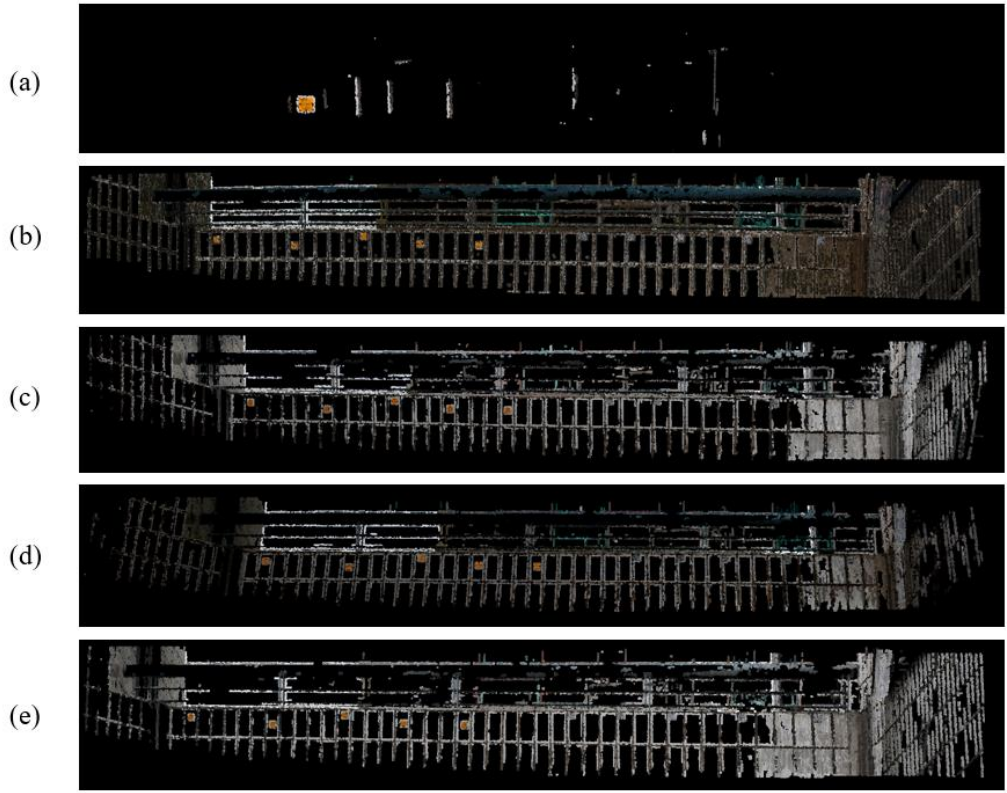

Figure 14. Point clouds of survey experiment 2 reconstructed from: (a) original images, (b) images enhanced by DCP, (c) images enhanced by AMEF, (d) images enhanced by AOD-Net, (e) images enhanced by proposed new method.

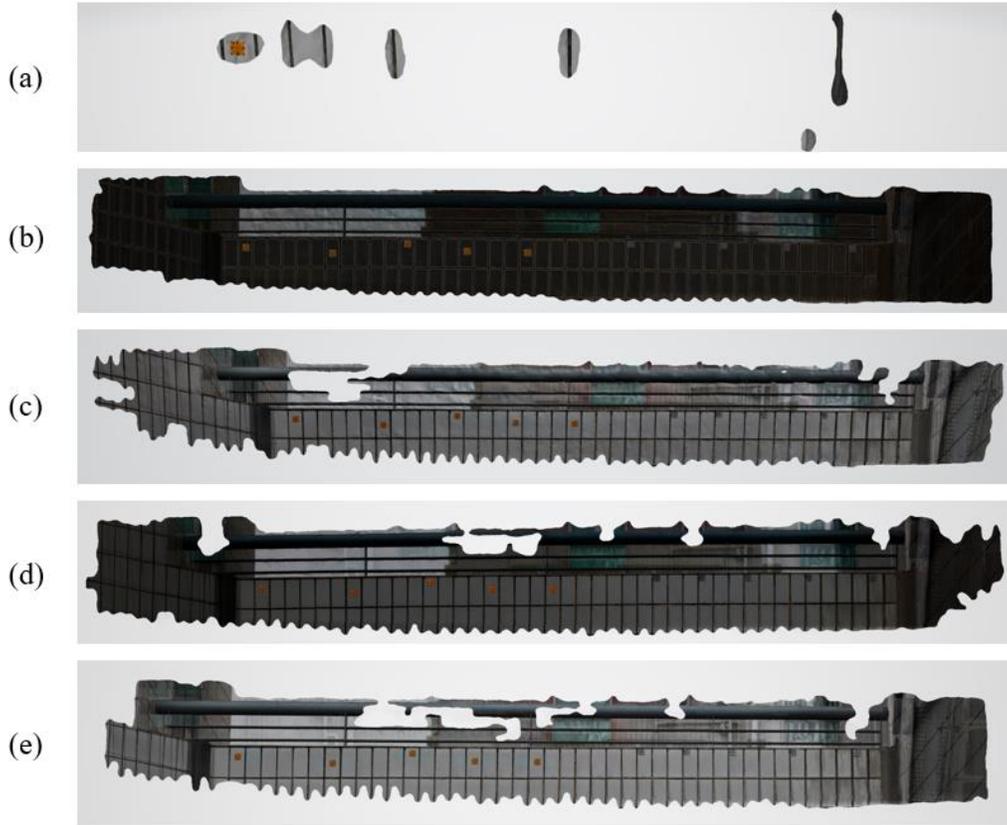

Figure 15. 3D mesh models (front views) of survey experiment 2 reconstructed from: (a) original images, (b) images enhanced by DCP, (c) images enhanced by AMEF, (d) images enhanced by AOD-Net, (e) images enhanced by proposed new method.

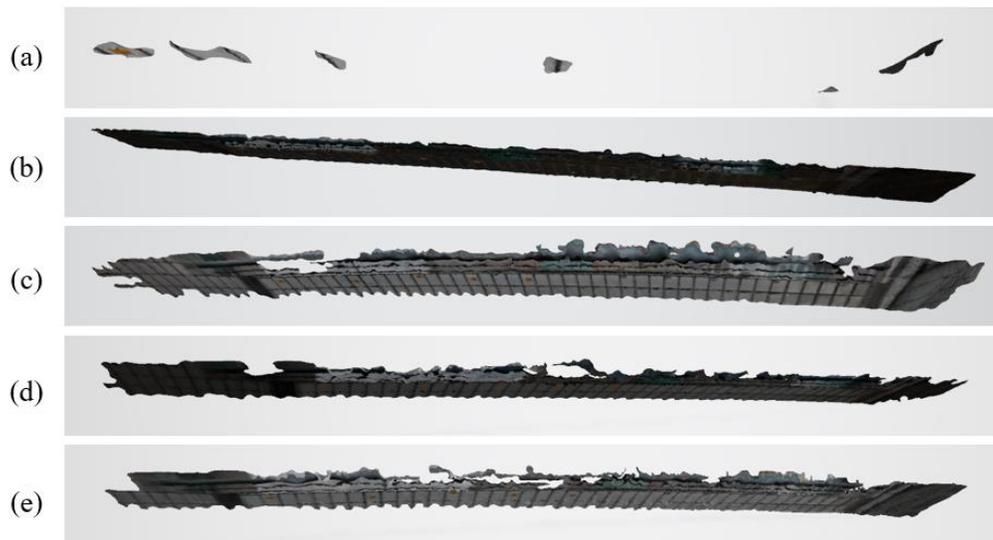

Figure 16. 3D mesh models (top views) of survey experiment 2 reconstructed from: (a) original images, (b) images enhanced by DCP, (c) images enhanced by AMEF, (d) images enhanced by AOD-Net, (e) images enhanced by proposed new method.

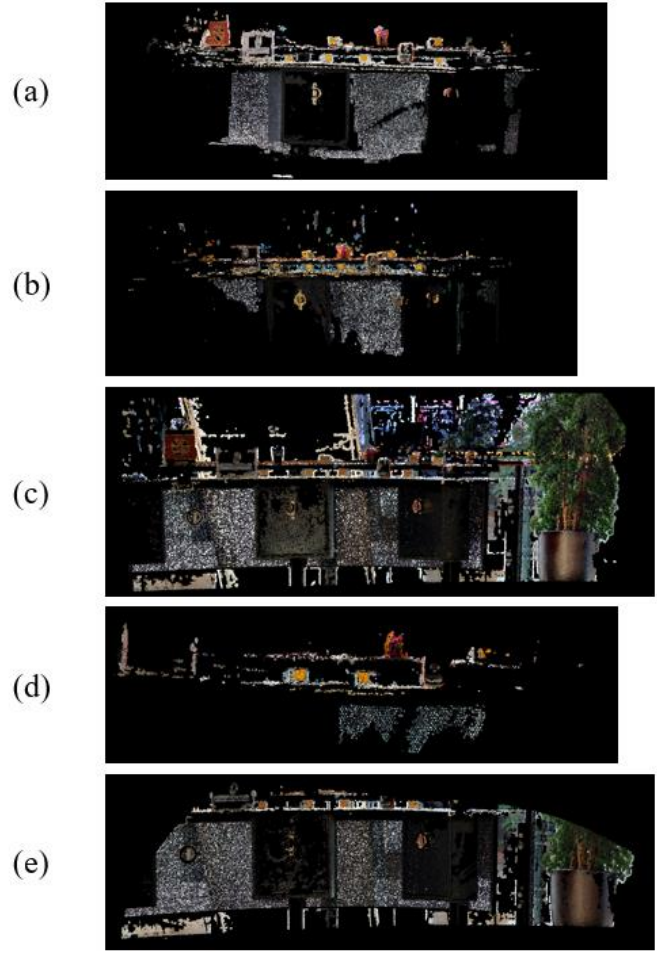

(a)

(b)

(c)

(d)

(e)

Figure 17. Point clouds of survey experiment 3 reconstructed from: (a) original images, (b) images enhanced by DCP, (c) images enhanced by AMEF, (d) images enhanced by AOD-Net, (e) images enhanced by proposed new method.

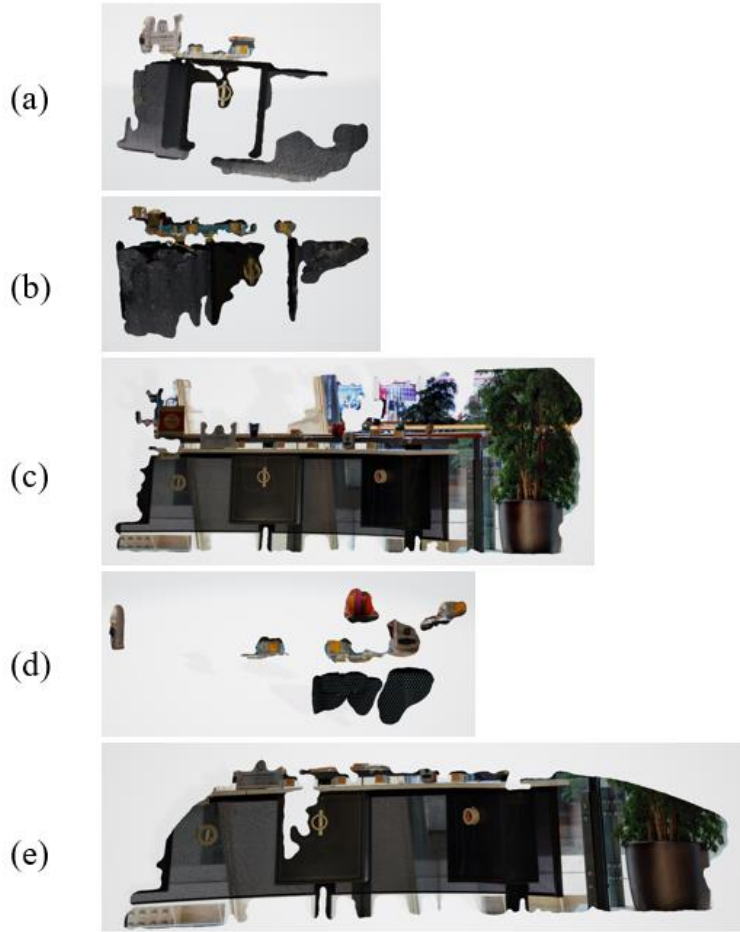

Figure 18. 3D mesh models (front views) of survey experiment 3 reconstructed from: (a) original images, (b) images enhanced by DCP, (c) images enhanced by AMEF, (d) images enhanced by AOD-Net, (e) images enhanced by proposed new method.

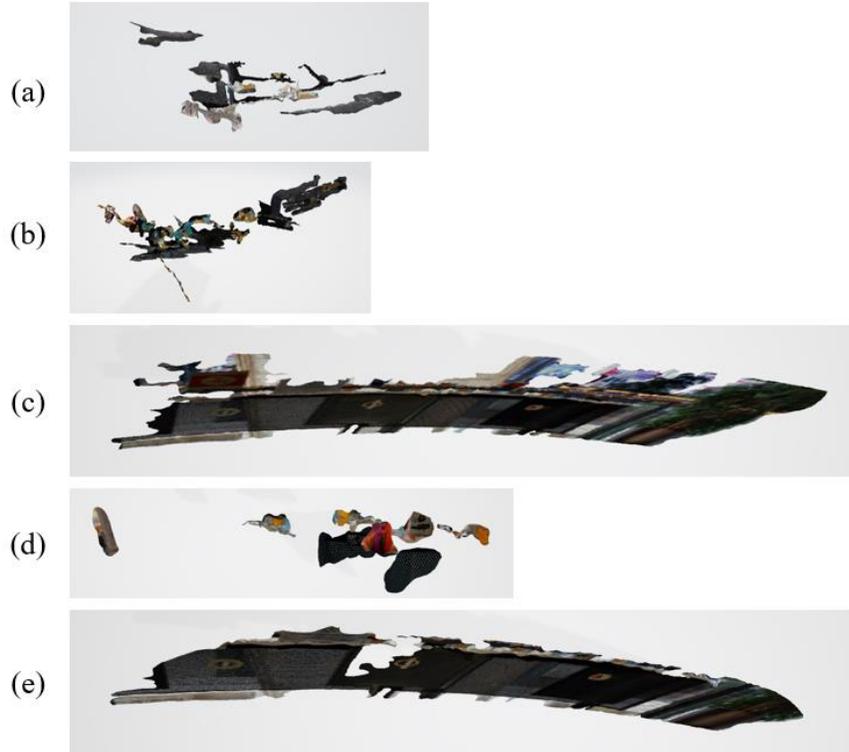

Figure 19. 3D mesh models (top views) of survey experiment 3 reconstructed from: (a) original images, (b) images enhanced by DCP, (c) images enhanced by AMEF, (d) images enhanced by AOD-Net, (e) images enhanced by proposed new method.

In summary, the original images produce 3D models that are inferior to those generated by enhanced images. For seriously degraded images, the photogrammetry software may even fail to reconstruct the 3D model. DCP and AOD-Net can reconstruct fairly accurate surface models. However, the texture map is much darker than the original images which means that the reconstructed model is far from photorealistic. Also, they may pro-duce very inaccurate 3D model with seriously degraded images. AMEF and proposed new method generate the most accurate and visually pleasing 3D models. By looking at the visual result in close detail, our method produces 3D models that are slightly better than AMEF. This difference is more obvious in the numerical results. The proposed new method always achieves lower mean errors than AMEF in all survey experiments.

## 6. Conclusion

We propose a method that can be used to enhance the contrast and visibility of survey images. The images, when fed to the photogrammetry software, can reconstruct a highly accurate 3D scene model with sub-millimeter errors. The image enhancement method, inspired by the idea of image dehazing, is first to transform the original degraded image into multiple exposure images. The local texture and saturation of the transformed images are analyzed and results in weight maps, which are used to synthesize the enhanced image. Performance evaluation has been performed on datasets with synthetic and real hazy images. Survey experiments are performed in both outdoor and indoor scenes. The proposed new method outperforms other well-known and state-of-the-art image dehazing methods. The images, when enhanced by the proposed new method, can reconstruct highly accurate 3D scene models.

In the future, more surveys will be performed. The image acquisition process can be automated with the use of a motorized pan-tilt head, remote control and wireless image file transmitter for the digital SLR camera. With the present camera and lens, it is not possible to capture a very clear image at a distance of more than 70 m. For monitoring at a longer distance, a lens with longer or fixed focal length can be used. The number of images acquired can be increased in order to increase the overlap between adjacent views. Moreover, we will continue our research on image enhancement such as image dehazing. Haze removal has attracted research interest as it can substantially improve the performance of various computer vision applications. Better algorithms will be proposed for high accuracy 3D modeling of survey scene.

**Acknowledgments**

The work described in this paper was partially supported by a grant from the Research Grants Council of the Hong Kong Special Administrative Region, China (Project No. CityU 11202319) and Hong Kong Innovation and Technology Commission and City University of Hong Kong (Project No. 9042823).

**References**

1. X. Liu, C. Zhao, Q. Zhang, C. Yang, J. Zhang. Characterizing and monitoring ground settlement of marine reclamation land of Xiamen New Airport, China with Sentinel-1 SAR datasets. Remote Sensing, 2019, Vol. 11, Issue 5.
2. M. Batur, O. Yilmaz, H. Ozener. A case study of deformation measurements of Istanbul Land Walls via terrestrial laser scanning. IEEE Journal of Selected Topics in Applied Earth Observations and Remote Sensing, 2020, Vol. 13, pp. 6362-6371.
3. Y. Ge, H. Tang, X. Gong, B. Zhao, Y. Lu, Y. Chen, Z. Lin, H. Chen, Y. Qin. Deformation monitoring of earth fissure hazards using terrestrial laser scanning. Sensors, 2019, Vol. 19, 1463.
4. S. Mikrut. Integration of digital images and laser scanning data for the survey of architectural objects. Image Processing & Communications, 2012, Vol. 17, No. 4, pp. 161-166.
5. C. Moses, D. Robinson, J. Barlow. Methods for measuring rock surface weathering and erosion: A critical review. Earth-Science Reviews, 2014, Vol. 135, pp. 141-161.
6. N.D. Cullen, A.K. Verma, M.C. Bourke. A comparison of structure from motion photogrammetry and traversing micro-erosion meter for measuring erosion on shore platforms. Earth Surface Dynamics, 2018, Vol. 6, pp. 1023-1039.
7. S.C. Chian, J. Yang. Use of photogrammetry for ground settlement measurement. Proceedings of the 19th International Conference on Soil Mechanics and Geotechnical Engineering, 2017, pp. 1663-1666.
8. V. Baiocchi, Q. Napoleoni, M. Tesei, G. Servodio, M. Alicandro, D. Costantino. UAV for monitoring the settlement of a landfill. European Journal of Remote Sensing, 2019, Vol. 52, No. S3, pp. 41-52.
9. P. Łabędź, K. Skabek, P. Ozimek, M. Nytko. Histogram adjustment of images for improving photogrammetry reconstruction. Sensors, 2021, Vol. 21, 4654.
10. G.H. Babu, N. Venkatram. A survey on analysis and implementation of state-of-the-art haze removal techniques. Journal of Visual Communication and Image Representation, 2020, Vol. 72, 102912.

11. A.M. Chaudhry, M.M. Riaz, A. Ghafoor. A framework for outdoor RGB image enhancement and dehazing. IEEE Geoscience and Remote Sensing Letters, 2018, Vol. 15, No. 6, pp. 932-936.

12. Q. Guo, H.-M. Hu, B. Li. Haze and thin cloud removal using elliptical boundary prior for remote sensing image. IEEE Transactions on Geoscience and Remote Sensing, 2019, Vol. 57, No. 11, pp. 9124-9137.

13. J. Li, Q. Hu, M. Ai. Haze and thin cloud removal via sphere model improved dark channel prior. IEEE Geoscience and Remote Sensing Letters, 2019, Vol. 16, No. 3, pp. 472-476.

14. S. Salazar-Colores, J.-M. Ramos-Arreguin, C.J.O. Echeverri, E. Cabal-Yepez, J.-C. Pedraza-Ortega, J. Rodriguez-Resendiz. Image dehazing using morphological opening, dilation and Gaussian filtering. Signal, Image and Video Processing, 2018, Vol. 12, pp. 1329-1335.

15. L. Zhang, S. Wang, X. Wang. Saliency-based dark channel prior model for single image haze removal. IET Image Processing, 2018, Vol. 12, Issue 6, pp. 1049-1055.

16. F.A. Dharejo, Y. Zhou, F. Deeba, Y. Du. A color enhancement scene estimation approach for single image haze removal. IEEE Geoscience and Remote Sensing Letters, 2020, Vol. 17, No. 9, pp. 1613-1617.

17. J. Liu, S. Wang, X. Wang, M. Ju, D. Zhang. A review of remote sensing image dehazing. Sensors, 2021, Vol. 21, No. 11, 3926.

18. W. Wang, X. Yuan. Recent advances in image dehazing. IEEE/CAA Journal of Automatica Sinica, 2017, Vol. 4, No. 3, pp. 410-436.

19. K. He, J. Sun, X. Tang. Single image haze removal using dark channel prior. Proceedings of IEEE Conference on Computer Vision and Pattern Recognition, 2009, pp. 1956-1963.

20. J. Xiao, L. Zhu, Y. Zhang, E. Liu, J. Lei. Scene-aware image dehazing based on sky-segmented dark channel prior. IET Image Processing, 2017, Vol. 11, Issue 12, pp. 1163-1171.

21. Y. Li, S. You, M.S. Brown, R.T. Tan. Haze visibility enhancement: a survey and quantitative benchmarking. Computer Vision and Image Understanding, 2017, Vol. 165, pp. 1-16.

22. A. Galdran. Image dehazing by artificial multiple-exposure image fusion. Signal Processing, 2018, Vol. 149, pp. 135-147.

23. B. Li, X. Peng, Z. Wang, J. Xu, D. Feng. AOD-Net: All-in-One Dehazing Network. Proceedings of International Conference on Computer Vision, 2017, pp. 4780-4788.

24. L. Jiao, C. Hu, L. Huo, P. Tang. Guided-Pix2Pix: end-to-end inference and refinement network for image dehazing. IEEE Journal of Selected Topics in Applied Earth Observations and Remote Sensing, 2021, Vol. 14, pp. 3052-3069.

25. C.O. Ancuti, C Ancuti, R. Timofte, C.D. Vleeschouwer. I-HAZE: A dehazing benchmark with real hazy and haze-free indoor images. Lecture Notes in Computer Science, 2018, LNCS 11182, pp. 620-631.

26. J.-P. Tarel, N. Hautiere, A. Cord, D. Gruyer, H. Halmaoui. Improved visibility of road scene images under heterogeneous fog. Proceedings of IEEE Intelligent Vehicles Symposium, 2010, pp. 478–485.

27. J.-P. Tarel, N. Hautiere, L. Caraffa, A. Cord, H. Halmaoui, D. Gruyer. Vision enhancement in homogeneous and heterogeneous fog. IEEE Intelligent Transportation Systems Magazine, 2012, Vol. 4, No. 2, pp. 6–20.

28. S. Zhao, L. Zhang, S. Huang, Y Shen, S. Zhao. Dehazing evaluation: real-world benchmark datasets, criteria, and baselines. IEEE Transactions on Image Processing, 2020, Vol. 29, pp. 6947-6962.

29. N. Hautière, J.-P. Tarel, D. Aubert, É. Dumont. Blind contrast enhancement assessment by gradient ratioing at visible edges. Image Analysis & Stereology, 2008, Vol. 27, No. 2, pp. 87–95.

30. L.K. Choi, J. You, A.C. Bovik. Referenceless prediction of perceptual fog density and perceptual image defogging. IEEE Transactions on Image Processing, 2015, Vol. 24, No. 11, pp. 3888–3901.
31. Z. Wang, A.C. Bovik, H.R. Sheikh, E.P. Simoncelli. Image quality assessment: from error visibility to structural similarity. IEEE Transactions on Image Processing, 2004, Vol. 13, No. 4, pp. 600–612.
32. https://www.pix4d.com
33. S. Liao, G. Zhao, V. Kellokumpu, M. Pietikäinen, S.Z. Li. Modeling pixel process with scale invariant local patterns for background subtraction in complex scenes. Proceedings of IEEE Conference on Computer Vision and Pattern Recognition, 2010, pp. 1301-1306.
34. P.-L. St-Charles, G.-A. Bilodeau, R. Bergevin. SuBSENSE: a universal change detection method with local adaptive sensitivity. IEEE Transactions on Image Processing, 2015, Vol. 24, No. 1, pp. 359-373.